\documentclass{article}

\usepackage{PRIMEarxiv}

\usepackage[utf8]{inputenc} 
\usepackage[T1]{fontenc}    
\usepackage{url}            
\usepackage{booktabs}       
\usepackage{amsfonts}       
\usepackage{amssymb}
\usepackage{amsthm}
\usepackage{amsmath}
\usepackage{multirow}
\usepackage{nicefrac}       
\usepackage{microtype}      
\usepackage{lipsum}
\usepackage{fancyhdr}       
\usepackage{graphicx}       
\usepackage[hyperfootnotes=true]{hyperref}
\usepackage{tabularx}
\graphicspath{{media/}}     

\title{A variational autoencoder-based nonnegative matrix factorisation model for deep dictionary learning}

\author{
  Hong-Bo Xie \\
  ARC Centre of Excellence for Mathematical \& Statistical Frontiers \\
  Queensland University of Technology \\
  Brisbane\\
  \texttt{hongbo.xie@qut.edu.au} \\
   \And
  Caoyuan Li, Shuliang Wang \\
  School of Computer Science and Technology \\
  Beijing Institute of Technology \\
  Beijing\\
    \And
  Richard Yi Da Xu \\
  Faculty of Engineering and Information Technology \\
  University of Technology Sydney \\
  Sydney \\
    \And
  Kerrie Mengersen \\
  ARC Centre of Excellence for Mathematical \& Statistical Frontiers \\
  Queensland University of Technology \\
  Brisbane\\
}

\begin{document}
\maketitle

\begin{abstract}
Construction of dictionaries using nonnegative matrix factorisation (NMF) has extensive applications in signal processing and machine learning. With the advances in deep learning, training compact and robust dictionaries using deep neural networks, i.e., dictionaries of deep features, has been proposed. In this study, we propose a probabilistic generative model which employs a variational autoencoder (VAE) to perform nonnegative dictionary learning. In contrast to the existing VAE models, we cast the model under a statistical framework with latent variables obeying a Gamma distribution and design a new loss function to guarantee the nonnegative dictionaries. We adopt an acceptance-rejection sampling reparameterization trick to update the latent variables iteratively. We apply the dictionaries learned from VAE-NMF to two signal processing tasks, i.e., enhancement of speech and extraction of muscle synergies. Experimental results demonstrate that VAE-NMF performs better in learning the latent nonnegative dictionaries in comparison with state-of-the-art methods.
\end{abstract}


\section{Introduction}
Dictionary learning, constructing a dictionary consisting of atoms to represent a class of signals, has been an effective tool to solve a number of problems in machine learning and signal processing, for example, face recognition \cite{huang2016learning}, speech enhancement \cite{nie2018deep}, blind separation of hyperspectral image \cite{charles2011learning}, brain-machine interface \cite{xie2017decoding}, and neural information decoding \cite{bizzi2013neural}. Since negative dictionaries may contradict physical or physiological reality and lack intuitive meaning in most of these applications, dictionary learning using nonnegative matrix factorisation (NMF), which imposes the nonnegativity constraint, has been therefore a popular method. Essentially, NMF factorises a data matrix into the product of two nonnegative matrices with different properties, in which one is termed as the dictionary matrix, and the other termed as the activation matrix. With the learned dictionary, one can discover and properly understand the crucial causes underlying the observation.

In Lee and Seung’s seminal NMF paper \cite{lee1999learning,lee2000algorithms}, they presented the algorithms based on multiplicative update rules to minimise the cost functions of the Frobenius norm or KL-divergence. In addition to this standard NMF model, many variants of NMF have been proposed for various applications, such as sparse NMF \cite{hoyer2004non}, robust NMF \cite{kong2011robust}, and discriminative NMF \cite{weninger2014discriminative}. These methods take the sparsity of the activations, robustness to outliers, and the discrimination power of the dictionary into account, respectively. In the last decade, with the advances and breakthrough in deep learning techniques, deep neural networks (DNN) has been one of the most influential machine learning paradigms. The concept of deep dictionary learning (DDL) has been recently proposed where a shallow model is replaced by a DNN. Lemme et al. \cite{lemme2012online} presented an online learning scheme for non-negative sparse coding using an autoencoder neural networks. In order to enforce non-negative weights, they introduced an asymmetric, piecewise linear decay function. Hosseini Asl et al. \cite{hosseini2015deep} used a general autoencoder with trainable weights for both the hidden and the output layers to extend the model in \cite{lemme2012online}. In addition, they modified the loss function with an extra regularisation term to reduce the number of nonnegative weights of each layer in training. Similar deep dictionary learning frameworks can also be found in \cite{chorowski2014learning,nguyen2013learning}, where a multilayer perceptron and Boltzmann machine are employed as the underlying network architecture, respectively. In comparison with NMF, these deep dictionary learning architectures have shown more precise part-based feature representation, better generalisation ability for the trained model, and improved classification accuracy in the face and text recognition \cite{lemme2012online,hosseini2015deep,chorowski2014learning,nguyen2013learning}.

All of the DNNs mentioned above are deterministic models. A coherent theoretical framework for understanding, analysing, and synthesising their architectures has remained elusive. A variational autoencoder (VAE), a new deep generative probabilistic model, casts learning representations for high-dimensional distributions as a variational inference problem \cite{kingma2013auto}. Learning a VAE amounts to the optimisation of an objective, balancing the quality of samples that are autoencoded through a stochastic encoder-decoder pair while encouraging the latent space to follow a fixed Gaussian prior distribution \cite{ghosh2019variational}. In other words, a VAE can be viewed as a deterministic autoencoder where Gaussian noise is added to the decoder’s input \cite{ghosh2019variational}. With a principled statistical approximation, unsupervised learning manner, and effective training scheme, VAE has gained substantial attention for learning a probability distribution over complex data that cannot be represented by conventional linear models. In addition, in comparison with deterministic DNNs, VAE is very similar to NMF in terms of minimisation of the global cost function, choice of the scalar cost function and data representation, and interpretation of the underlying statistical model \cite{girin2019notes}. However, the original formulation of VAE \cite{kingma2013auto} assumes the latent variables obeys a Gaussian distribution, violating the requirement of nonnegativity. Our major contribution in this Letter is to develop a variational autoencoder-based nonnegative matrix factorisation (VAE-NMF) model for dictionary learning. We employ a Gamma distribution to replace the Gaussian distribution to represent the latent variables, which is consistent with the reality of many real-world data sets \cite{kwon2016nmf,moussaoui2006separation,ross2020introduction}. To this end, we propose a new VAE loss function which forces the weights to be nonnegative. With these two new model assumptions and learning schemes, both the dictionary and activation are therefore nonnegative. We test the proposed model on two benchmark data sets to enhance speech and extract muscle synergies, respectively.

Experimental results demonstrate that VAE-NMF achieves improved performance in comparison with representative baseline methods, including Lee and Seung’s standard NMF \cite{lee1999learning,lee2000algorithms}, the discriminative NMF \cite{weninger2014discriminative}, and a deterministic DNN \cite{huang2014deep}. The rest of this paper is organized as follows. In Section II, we briefly introduce the basic structure and principle of VAE for self-containment and elaborate on the details of model specification and inference for the VAE-NMF method. Results of two experimental signals, comparison with state-of-the-art methods, and objective assessments are presented in Section III. Finally, Section IV concludes the Letter.

\section{Methods}
\label{sec:headings}
\subsection{Problem Formulation}
For an observed matrix $\mathbf{X} \in \mathbb{R}^{m \times n}$ with column vectors $\mathbf{x} \in \mathbb{R}^{m \times 1}$, VAE defines a probabilistic generative model of data distribution:
\begin{equation}
p_{\theta}(\mathbf{x}, \mathbf{z}) = p_{\theta}(\mathbf{x}|\mathbf{z})p_{\theta}(\mathbf{z})
\end{equation}
where $\mathbf{z} \in \mathbb{R}^{r \times 1}$ is the latent stochastic variable with rank $r < min(m, n)$ in our VAE-NMF model. The prior $p_{\theta}(\mathbf{z})$ quantifies what we know about $\mathbf{z}$ independent of independent of the observed data, and it will work as a regulariser in the model. The likelihood function $p_{\theta}(\mathbf{x}|\mathbf{z})$ quantifies how the generation of data $\mathbf{x}$ is conditioned on the latent variable $\mathbf{z}$. Here both $p_{\theta}(\mathbf{z})$ and $p_{\theta}(\mathbf{x}|\mathbf{z})$ are parametric families of distributions with parameters $\theta$. A VAE consists of an encoder and a decoder. The encoder parameterised with $\theta$ infers $\mathbf{z}$ from $\mathbf{x}$. Since the exact posterior $p_{\theta}(\mathbf{z}|\mathbf{x})$ is usually intractable, we approximate it with some tractable auxiliary distribution $q_{\phi}(\mathbf{z}|\mathbf{x})$ with parameter set $\phi$. This part of the model parameterised with $\phi$ to generate $\mathbf{x}$ from $\mathbf{z}$ is termed the decoder. The task of training a VAE is to estimate the parameters $\theta$ and $\phi$ for fitting the approximate posterior to the true posterior. This can be achieved by optimising a lower bound of the marginal log-likelihood log $p_{theta}(\mathbf{x})$ estimated from the training dataset of the matrix $\mathbf{X}$. For a single column training sample $\mathbf{x}$, the corresponding marginal log-likelihood can be written as \cite{kingma2013auto}:
\begin{equation}
\log p_{\theta}(\mathbf{x}) = D_{KL}(q_{\phi}(\mathbf{z}|\mathbf{x})|| p_{\theta}(\mathbf{x}|\mathbf{z})) + \mathcal{L}(\phi, \theta, \mathbf{x})
\end{equation}
where $D_{KL}$ represents the non-negative Kullback-Leibler (KL) divergence and $\mathcal{G}(\phi, \theta, \mathbf{x})$ is the variational lower bound (VLB) denoted by
\begin{equation}
g(\phi, \theta, \mathbf{x}) = \mathbb{E}_{q_{\phi}(\mathbf{z}|\mathbf{x})}[\log p_{\theta}(\mathbf{x}|\mathbf{z})] - D_{KL}(q_{\phi}(\mathbf{z}|\mathbf{x})|| p_{\theta}(\mathbf{z}))
\end{equation}
In this objective function (3), the first term is the expectation of the log-likelihood that the input data can be generated based on the sampled values of $\mathbf{z}$ from the inferred distribution $q_{\phi}(\mathbf{z}|\mathbf{x})$. The second term is the KL divergence between the distribution of $\mathbf{z}$ inferred from $\mathbf{x}$ and the prior distribution of $\mathbf{z}$. In the original formulation of the seminal paper \cite{kingma2013auto}, all the related distributions here are assumed Gaussians, and the objective function is, therefore, differentiable with respect to $(\theta, \phi)$ as well as the mean and variance of $q_{\phi}(\mathbf{z}|\mathbf{x})$. The parameters in the VAE then could be iteratively updated using stochastic gradient-descent algorithms. However, NMF aims to decompose $\mathbf{X} \in \mathbb{R}_{+}^{m\times n}$ into the product of a dictionary $\mathbf{W} \in \mathbb{R}_{+}^{m\times r}$ and an activation coefficient $\mathbf{Z} \in \mathbb{R}_{+}^{r \times n}$. The original VAE model cannot guarantee nonnegative latent vector and weights.

\begin{figure}
\centerline{\includegraphics[width=\columnwidth]{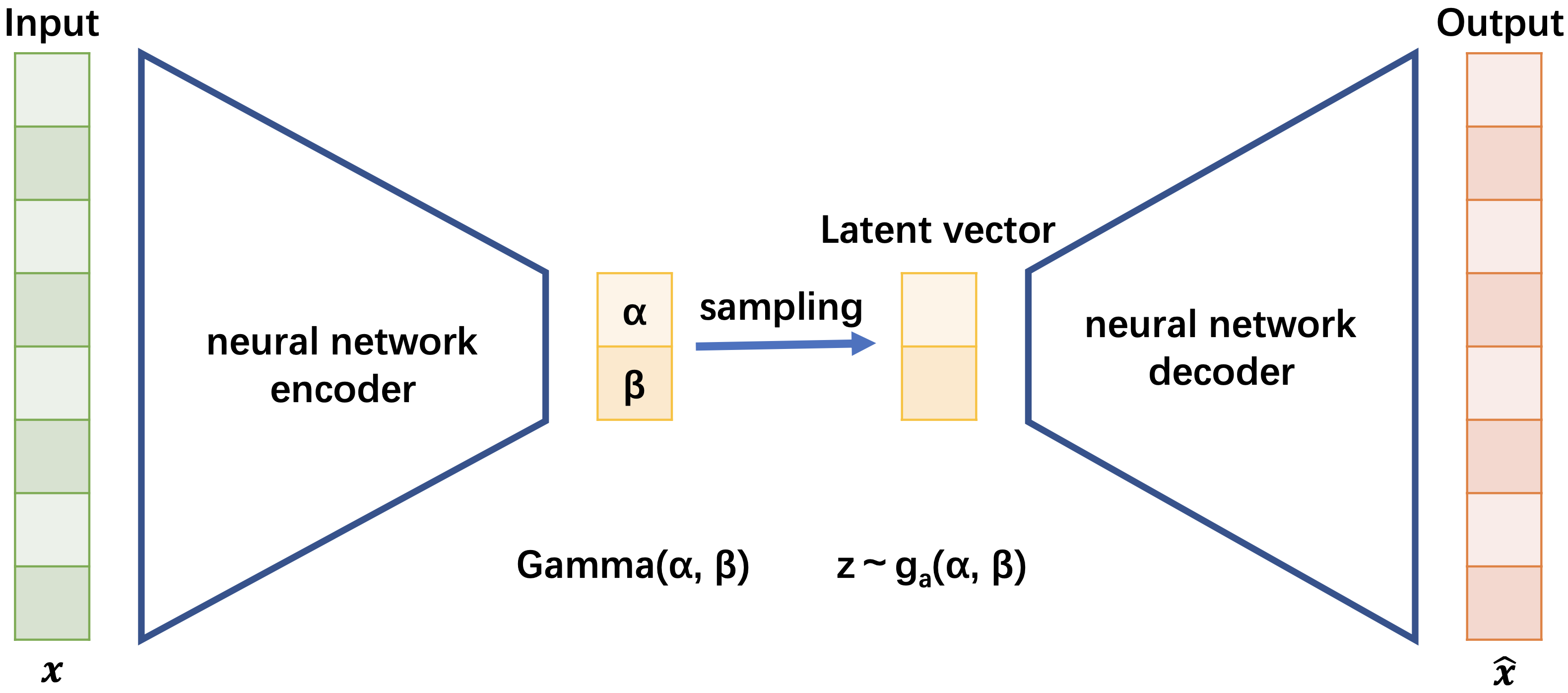}}
\caption{The schematic representation of VAE-NMF model.}
\label{fig:fig1}
\end{figure}

\subsection{VAE-NMF model}
Fig.\ref{fig:fig1} is the schematic representation of the proposed VAE-NMF model. The encoder consists of a deep-layer neural network. We set the decoder with a single latent layer variable, $\mathbf{Z} \in \mathbb{R}_{+}^{r \times n}$ and the corresponding weights are $\mathbf{W} \in \mathbb{R}_{+}^{m\times r}$. The output of the decoder is $\hat{\mathbf{X}} = \mathbf{WZ}$ to form an NMF structure. For this purpose, we need to modify both the loss function of the model and distributions related to the latent variable $\mathbf{z}$. Since the objective function we obtain in Eq. 3 is to be maximised during training, commonly it may be viewed as a “gain” function as opposed to a loss function. To obtain a loss function to promote nonnegative weights $\mathbf{W}$ of the decoder, we simply take the negative of $\mathbf{G}$ and add a regularisation term:
\begin{equation}
p(\mathbf{z}) = \mathcal{L}(\phi, \theta, \mathbf{x}) = -g(\phi, \theta, \mathbf{x}) + \frac{\gamma}{2}\sum_{i=1}^{r}\sum_{j=1}^N f(w_{ij}),
\end{equation}
where $\gamma > 0$ and $f(w_{i, j})$ is defined as a uadratic function \cite{hosseini2015deep}:
\begin{equation}
f(w_{ij}) = \left\{
\begin{array}{lr}
w^2_{i,j} & if\ w_{ij} < 0, \\
0 & if\ w_{ij} \geq 0.
\end{array}
\right.
\end{equation}

One can now find that minimisation of Eq. 4 results in reducing the average reconstruction error while pruning the nonnegative weights of the decoder. For the latent variable $\mathbf{z}$, we assume it obeys a Gamma distribution with shape parameter $\alpha$ and rate parameter $\beta = 1$:

\begin{equation}
p(\mathbf{z}) = \frac{\mathbf{z}e^{-\mathbf{z}}}{\Gamma(\alpha)},
\end{equation}

where $\Gamma (\alpha)$ is the gamma function. Similar to using a standard Gaussian distribution in Gaussian VAE, we can always reparameterize the rate if $\beta \neq 1$. It should be noted that as the shape parameter $\alpha$ becomes large, the Gamma distribution starts to resemble the Gaussian distribution, which can be theoretically explained by the central limit theorem \cite{ross2020introduction}. More importantly, many real-world data, for example, speech spectrogram and physiological signals, can be better fitted by using a Gamma distribution \cite{moussaoui2006separation,saji2015probability}. To sample $\mathbf{z}$ from a Gamma distribution, the explicit reparameterization gradients in original VAE is not applicable since the analytic form of the inverse cumulative function of the Gamma distribution is not available. We resort to the acceptance-rejection sampling reparameterization trick to sample $\mathbf{z}$ with \cite{naesseth2017reparameterization}
\begin{equation}
\mathbf{z} = h(\varepsilon, \alpha) = (\alpha - \frac{1}{3})(1+\frac{\varepsilon}{\sqrt{9\alpha-3}})^3,
\end{equation}
where $\varepsilon \sim \mathcal{N}(0, 1)$. We divide $\mathbf{z}$ by the rate $\beta$ to obtain a sample distributed as $Gamma(\alpha, \beta)$. So far, we have all quantities at hand except the analytical form of the KL divergence for differentiating the loss function. Following \cite{bauckhage2014computing}, the KL divergence between two Gamma distributions is given by:
\begin{equation}
\begin{split}
D_{KL}(f_1||f_2) &= \int_0^{\infty} f_1(x_1|\alpha_1, \beta_1)\log(\frac{f_1(x|\alpha_1, \beta_1)}{f_2(x_2|\alpha_2, \beta_2)})dx\\
&=\log\frac{\beta_1^{\alpha_1} \Gamma(\alpha_2)}{\beta_2^{\alpha_2} \Gamma(\alpha_1)} + [\Psi(\alpha_1)-\log\beta_1)](\alpha_1-\alpha_2) + \alpha_1(\frac{\beta_1}{\beta_2}-1).
\end{split}
\end{equation}

\section{Results}
\begin{figure}
\centerline{\includegraphics[width=\columnwidth]{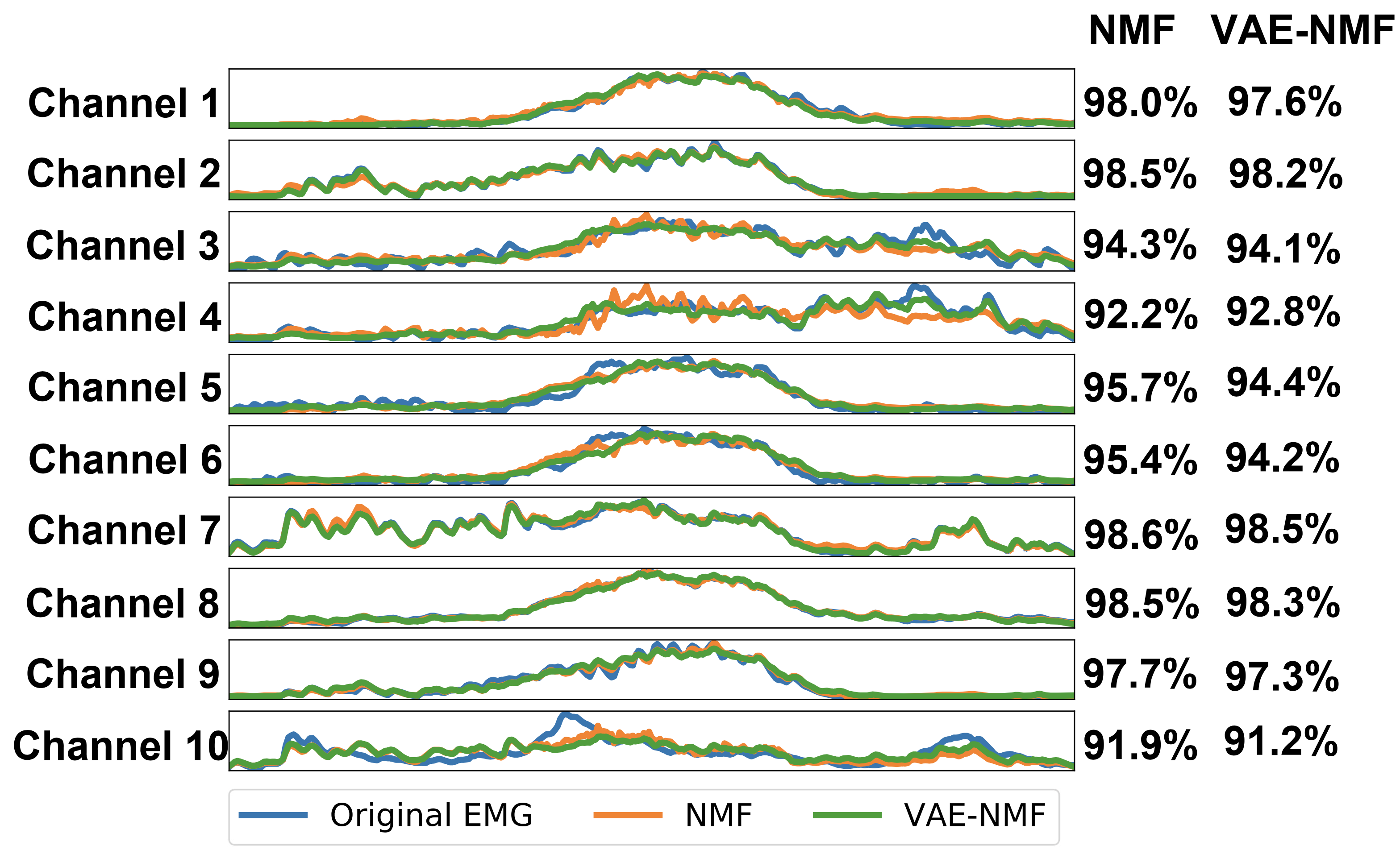}}
\caption{The original $10$-channel EMGs and the reconstructed EMGs by VAE-NMF and the standard NMF}
\label{fig:fig2}
\end{figure}

We test the performance of VAE-NMF on two publicly available data sets. The encoder neural network consists of $2$ hidden layers with $400$ nodes each. When training the neural network, the batch size is set to $128$. We utilized the Adam optimizer with the learning rate set as $10^{-3}$ and weight decay parameter set as $5\times 10^{-4}$. We first apply VAE-NMF as a dictionary learning-based data generator. In neuroscience, it is supposed that the central neural system controls muscle synergies, or groups of co-activated muscles, rather than individual muscles, to organise any simple or complex actions and movements. Muscle synergies, i.e. the nonnegative dictionary $\mathbf{W}$, extracted from multichannel EMG signals using NMF have been widely applied in human-machine interfaces, prosthetic controls, neural system disease diagnoses, and stroke rehabilitation \cite{atzori2014characterization}. The first Ninapro dataset consists of $10$-channel EMG recordings for wrist, hand and finger movements \cite{atzori2014characterization}. Each movement/task has 10 repetitions. The learned dictionary (synergies) are then used to reconstruct the muscle activities. The Variance Accounted For (VAF) is the major index to quantify how well the extracted synergies approximates the recorded EMG activity \cite{delis2015task,cheng2019motor,roh2013alterations}. The standard NMF method tends to overfit the EMG data with high VAF values. For both VAE-NMF and standard NMF techniques, we estimate the synergies from training repetitions for each task independently. We follow the previous recommendation to set the rank of the dictionary, i.e., the dimension of the latent vector $\mathbf{z}$, as four \cite{delis2015task,cheng2019motor,roh2013alterations}. Fig.\ref{fig:fig2} shows the original EMG signals as well as reconstructions by VAE-NMF and NMF. The VAF value of both methods for each muscle is also indicated in Fig.\ref{fig:fig2}. We find that both methods can reconstruct the EMG activity with VAF over $90\%$. However, for all muscles, the VAF values obtained from VAE-NMF are lower than that of NMF, which means the former avoids or at least alleviates overfitting. This can be explained by the fact that the KL divergence term is imposed in the loss function and the latent outputs are sampled from the distribution in VAE-NMF rather than taken deterministically in standard NMF. These two impositions help to reduce overfitting which results in higher VAF values.

We then train the dictionaries of clean speech on TIMIT and noise on NOISEX-$92$ databases, respectively, for speech enhancement. The factory and babble noise is used in the experiment. The dictionary of clean speech $\mathbf{W}^{S}$ is trained based on a $10000$-frame spectrogram of the clean speech data while the noise dictionary $\mathbf{W}^n$ is trained by using a $9000$-frame spectrogram of noise data. The rank of speech and noise dictionaries is set to be $40$ each. To test the performance of the learned dictionaries, a set of noisy speech utterances is synthesised with SNRs from $-5$ to $20$ dB with steps of $5$ dB. We then follow the procedure in \cite{kang2014nmf} to construct input noisy data vector and output encoding vector pairs to train another DNN using $\mathbf{W}^s$ and $\mathbf{W}^n$. Readers are referred to \cite{kang2014nmf} for the details and the related code are available online\footnote{\href{https://github.com/eesungkim/Speech_Enhancement_DNN_ NMF/}{https://github.com/eesungkim/Speech\_Enhancement\_DNN\_NMF/}}. We train VAE-NMF, as well as standard NMF \cite{hoyer2004non}, DNMF \cite{weninger2014discriminative} and recurrent neural network \cite{huang2014deep} models for comparison. The performance of the speech enhancement is evaluated in terms of the signal to distortion ratio (SDR), signal to interference ratio (SIR), signal to artefacts ratio (SAR) \cite{delis2015task} and the perceptual evaluation of speech quality (PESQ) score \cite{cheng2019motor}. Table I shows the SDR, SIR, and SAR values and PESQ scores for enhanced speech obtained from four algorithms averaged over two noise types. One can find that the VAE-NMF model is superior to the conventional NMF-based dictionary learning techniques and the recurrent neural network. We further test the performance of VAE-NMF and competitive models when they are trained over two types of noise pooled together. Table \ref{tab:table2} shows the scores of four methods. Compared with Table \ref{tab:table1}, the performance of all methods declines due to the noise dictionary trained over the pooled noise. However, the two deep learning-based methods have less performance decline, while VAE-NMF still outperforms three other methods in terms of almost all four scores over different SNRs. This demonstrates that both deep dictionary and deep neural network contribute to the speech enhancement in this test.

\begin{table}
 \caption{Speech enhancement performance with four dictionary learning algorithms over models separately trained models for factory and babble noise}
  \centering
  \resizebox{\linewidth}{!}{
  \begin{tabular}{|c|c|c|c|c|c|c|c|c|c|c|c|c|c|c|c|c|}
    \hline
   \multirow{2}*{SNR (dB)}     & \multicolumn{4}{|c|}{NMF}     & \multicolumn{4}{|c|}{DNMF} & \multicolumn{4}{|c|}{DNN} & \multicolumn{4}{|c|}{VAE-NMF} \\
    \cline{2-17}
    ~ & SDR & SIR & SAR & PESQ & SDR & SIR & SAR & PESQ & SDR & SIR & SAR & PESQ & SDR & SIR & SAR & PESQ \\
    \hline
    -5 & 2.90 & 5.65 & 8.62 & 1.98 & 4.62 & 8.56 & 8.85 & 2.03 & 5.83 & 10.49 & 8.41 & 2.03 & 5.85 & 10.61 & 8.57 & 2.05 \\
    \hline
    0 & 7.15 & 10.17 & 11.56 & 2.33 & 8.28 & 11.66 & 12.15 & 2.42 & 9.68 & 13.92 & 12.31 & 2.56 & 9.73 & 13.96 & 12.40 & 2.42 \\
     \hline
    5 & 11.14 & 15.38 & 13.63 & 2.72 & 11.87 & 15.55 & 14.92 & 2.73 & 13.12 & 17.45 & 15.41 & 2.84 & 13.37 & 17.52 & 15.48 & 2.85 \\
     \hline
    10 & 14.25 & 19.62 & 15.82 & 2.96 & 15.08 & 18.82 & 17.99 & 2.99 & 16.76 & 19.14 & 19.30 & 3.21 & 16.94 & 19.57 & 19.36 & 3.16 \\
     \hline
    Aver & 8.86 & 12.71 & 12.41 & 2.50 & 9.96 & 13.65 & 13.48 & 2.54 & 11.34 & 15.25 & 13.86 & 2.66 & 11.47 & 15.42 & 13.95 & 2.62 \\
    \hline
  \end{tabular}}
  \label{tab:table1}
\end{table}

\begin{table}
 \caption{Speech enhancement performance with four dictionary learning algorithms over model trained with pooled factory and babble noise}
  \centering
  \resizebox{\linewidth}{!}{
  \begin{tabular}{|c|c|c|c|c|c|c|c|c|c|c|c|c|c|c|c|c|}
    \hline
   \multirow{2}*{SNR (dB)}     & \multicolumn{4}{|c|}{NMF}     & \multicolumn{4}{|c|}{DNMF} & \multicolumn{4}{|c|}{DNN} & \multicolumn{4}{|c|}{VAE-NMF} \\
    \cline{2-17}
    ~ & SDR & SIR & SAR & PESQ & SDR & SIR & SAR & PESQ & SDR & SIR & SAR & PESQ & SDR & SIR & SAR & PESQ \\
    \hline
    -5 & 1.95 & 5.36 & 7.10 & 1.89 & 3.94 & 7.46 & 8.74 & 1.96 & 5.45 & 9.57 & 7.52 & 2.12 & 5.73 & 11.71 & 7.80 & 2.00 \\
    \hline
    0 & 6.02 & 9.91 & 9.87 & 2.13 & 7.72 & 11.01 & 11.83 & 2.32 & 9.47 & 13.40 & 12.21 & 2.52 & 9.65 & 14.32 & 11.66 & 2.42 \\
     \hline
    5 & 9.64 & 14.90 & 11.93 & 2.52 & 11.35 & 14.86 & 14.65 & 2.62 & 13.23 & 17.16 & 15.07 & 2.81 & 13.27 & 18.16 & 15.15 & 2.83 \\
     \hline
    10 & 12.39 & 19.37 & 13.82 & 2.85 & 14.60 & 18.45 & 17.32 & 2.93 & 16.84 & 20.63 & 18.39 & 3.04 & 16.82 & 21.41 & 18.21 & 3.12 \\
     \hline
    Aver & 7.50 & 12.39 & 10.68 & 2.35 & 9.40 & 12.95 & 13.14 & 2.46 & 11.25 & 15.19 & 13.30 & 2.62 & 11.37 & 16.40 & 13.21 & 2.59 \\
    \hline
  \end{tabular}}
  \label{tab:table2}
\end{table}

\section{Conclusion}

We have presented a new nonnegative matrix factorization model for deep dictionary leaning using a variational autoencoder. To this end, we propose a new VAE loss function and perform a Gamma distribution-based reparameterization trick to sample the latent variables. Experiments on muscle synergies extraction and speech enhancement demonstrate its superiority to both conventional NMF dictionary learning methods and the recurrent neural network. The proposed VAE-NMF framework is applicable to analyse other real-world nonnegative data sets with truncated Gaussian, general Gamma, and exponential distributions.

\bibliographystyle{unsrt}  
\bibliography{references}

\end{document}